\documentclass[lettersize,journal]{IEEEtran}
\usepackage{amsmath}
\usepackage{amsfonts}
\usepackage{algorithm}
\usepackage{array}
\usepackage[caption=false,font=normalsize,labelfont=sf,textfont=sf]{subfig}
\usepackage{textcomp}
\usepackage{stfloats}
\usepackage{url}
\usepackage{verbatim}
\usepackage{graphicx}
\usepackage{cite}
\usepackage{multirow}
\usepackage{graphicx}
\usepackage{epstopdf}
\usepackage{color}   
\definecolor{blush}{rgb}{0.62,0.24,0.81}
\definecolor{royalazure}{rgb}{0.0,0.22,0.66}
\definecolor{uared}{rgb}{0.85,0.0,0.3}
\usepackage{diagbox}
\usepackage{tablefootnote}
\usepackage{threeparttable}
\everymath{\displaystyle}
\definecolor{black}{rgb}{0,0,0}

\begin{document}


\title{A Geometry-Driven, Framework-Agnostic Optimization for Object Pose Estimation}

\author{Wei Chen, Tao Zhen, Jing Zhang,  Zhongchen Shi, Liang Xie, Erwei Yin
}

\markboth{Journal of \LaTeX\ Class Files,~Vol.~xx, No.~x, xxx~xxx}%
{Shell \MakeLowercase{\textit{et al.}}: A Sample Article Using IEEEtran.cls for IEEE Journals}


\maketitle

\begin{abstract}
Current object pose estimation research remains predominantly model-centric, focusing on architectural innovations and post-processing refinements. This paper introduces a data-centric optimization by proposing a novel, physically grounded rotation representation through principal axes alignment. Our method aligns the object's coordinate system with its inherent geometric axes, derived from inertial properties, yielding three key advantages: 
Inherent Stability—leveraging the energy-minimizing property of principal axes provides a robust representation that is less sensitive to noise and occlusions; Symmetry-Aware Canonicalization—explicitly resolving rotational ambiguities for symmetric objects at the data level, which fundamentally eliminates label confusion during network training; and Framework Agnosticism—the optimization is applied purely at the dataset level, ensuring plug-and-play compatibility with existing networks without any architectural modification. We validate the framework across diverse category-level and instance-level models. Extensive experiments demonstrate consistent and significant accuracy improvements, while preserving the integrity of the baseline network. This work establishes a new, geometry-driven direction for enhancing pose estimation, circumventing the need for complex network redesign. 
\end{abstract}


\section{Introduction}
\IEEEPARstart{E}{stimating} the 6 Degree-of-Freedom (6D) pose—comprising an object's 3D orientation (rotation) and 3D translation—is a foundational computer vision task essential for bridging perceptual understanding with physical interaction. Robust and accurate 6D pose estimation underpins critical applications such as augmented reality (requiring precise virtual-real alignment) \cite{marchand2016pose}, autonomous driving (relying on exact localization of obstacles and agents), and robotic manipulation (demanding real-time, millimeter-accurate control of tools and workpieces) \cite{zhu2014single,tremblay2018deep}. 

Significant progress has been made in 6D object pose estimation for both instance-level and category-level tasks, as evidenced by extensive recent work \cite{Xiang2017,he2020pvn3d, Chen_2021_CVPR, Weng_2021_ICCV, chen2021sgpa, Li_2025_CVPR, lin2024instance, Huang_2024_CVPR, Wen_2024_CVPR}. 
However, the prevailing research focus has been on improving neural networks themselves, primarily through three types: (1) designing increasingly complex architectures with multi-stage pipelines, such as attention modules, and sophisticated feature fusion mechanisms \cite{he2020pvn3d, Chen_2020_WACV, Chen_2021_CVPR}; (2) developing specialized loss functions to address challenges such as object symmetries, and non-linear error metrics \cite{Liu_2023_ICCV, Hsiao_2024_CVPR, Zhao_2023_ICCV, Mo_2022_CVPR}; and (3) exploring diverse rotation representations, including quaternions, axis-angle, and continuous embeddings, to facilitate more accurate regression \cite{Xiang2017, zhou2019continuity}.

While these network-centric advancements have contributed to higher accuracy, they introduce notable practical limitations. First, integrating such enhancements often entails considerable implementation complexity. For instance, the adoption of specialized rotation representations \cite{zhou2019continuity} not only requires algorithmic adaptations but may also be incompatible with frameworks that do not directly regress rotation parameters, such as keypoint-based \cite{peng2018pvnet} or correspondence-driven methods \cite{Chen_2021_CVPR}. 
Second, post-refinement modules \cite{Li2018, Wen_2024_CVPR} are conceptually straightforward. However, they inevitably increase inference latency. This added computational overhead can compromise real-time performance, making such approaches less suitable for latency-sensitive applications, such as robotic manipulation or interactive augmented reality.
Third, for symmetric objects, existing methods either rely on handcrafted primitives or loss functions, or employ complex architectures to achieve an approximate mapping; neither of these approaches provides a principled solution for resolving rotational ambiguity.


\begin{figure}[h]
\centering
\includegraphics[width=0.45\textwidth]{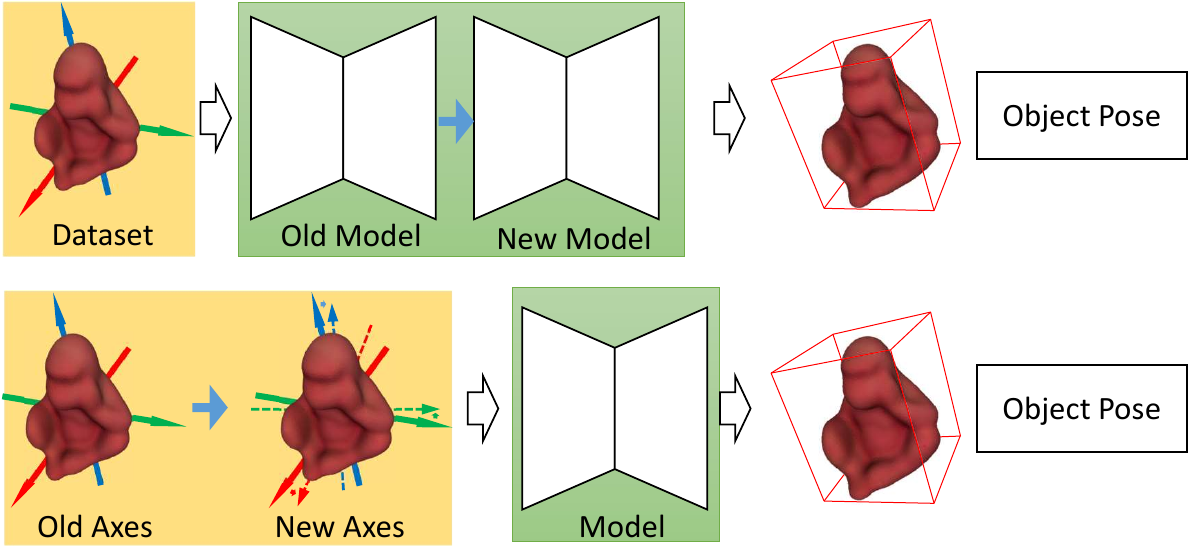}
\caption{\textbf{Semantic illustration of proposed mechanism}.The main contribution of previous methods mainly focused on network layout update, including new loss functions or new network architectures. 
Our proposed method only updates the rotation coordinate for pose representation, which is easy to apply and perfectly compatible with existing methods.}
\label{Fig:over_contr}
\end{figure}

To overcome the aforementioned issues, in this paper, we design a simple mechanism that improves the pose estimation accuracy without changing any layout details of the existing network or increasing the running time.

The proposed method, termed the $\pi$-Mechanism (Principal Axes of Inertia, PAI), establishes a canonical object coordinate system by aligning it with the three principal axes of inertia derived from the object's 3D geometry. This alignment is achieved by calculating the rotation matrix that maps the original object coordinates to this new, intrinsic reference frame.
Consequently, all ground-truth pose annotations in the training set are transformed into this stable coordinate system via this matrix. A network is then trained or fine-tuned to regress poses relative to this canonical frame.
The key advantages of this approach are threefold:
1) It is model-agnostic, as it operates purely on the dataset without altering the network architecture;2) it is computationally lightweight, requiring only a single matrix multiplication per object during pre-processing; 3) it resolves the ambiguous rotations of symmetric objects in a simple and analytical way.

To summarize, this paper makes the following primary contributions to the field of 6D object pose estimation:
\begin{itemize}
\item \textbf{A Data-Centric, Geometry-Driven Framework:} We introduce a paradigm shift from modifying network architectures to optimizing their geometric input. The core of this framework is the proposed $\pi$-Mechanism, a plug-and-play, model-agnostic module that canonicalizes an object's coordinate system by aligning it with its intrinsic Principal Axes. 
This establishes a stable reference frame rooted in classical mechanics, which is applied through a simple, one-time data preprocessing step requiring no changes to existing network architectures, loss functions, or training pipelines.

\item \textbf{Symmetry-Aware Canonicalization via Inertial Principal Axes:} We propose a data-level canonicalization scheme that explicitly resolves rotational ambiguities for symmetric objects. By aligning objects to their inherent principal axes and disambiguating their orientations with deterministic rules, our method eliminates label confusion at the source, leading to significantly improved pose estimation accuracy on symmetric objects without any network modification.

\item \textbf{Extensive and Conclusive Empirical Validation:} We demonstrate consistent and significant accuracy improvements across multiple benchmarks (LINEMOD, LINEMOD-OCC, NOCS-REAL), covering both instance-level and category-level tasks under various challenges. The results robustly validate that the introduced geometric stability directly translates to enhanced learning performance and generalization.

\end{itemize}

\section{Related Work}
\subsection{Network Architecture Optimization}
Recent advances in 6D pose estimation have focused on enhancing network architectures through multimodal fusion, computational efficiency, and cross-domain generalization. Early approaches, such as PoseCNN \cite{Xiang2017}, relied primarily on RGB input followed by geometric post-processing. Subsequent work emphasized the fusion of multimodal data, as seen in DenseFusion \cite{wang2019densefusion}, which performs pixel-wise fusion of RGB and point cloud features. More recent architectures have incorporated transformer-based modules for cross-modal attention, exemplified by Trans6D \cite{zhang_trans6d_2023}. To improve feature extraction under occlusion, GatedUniPose \cite{feng2024gateduniposenovelapproachpose} integrated gated convolutions with UniRepLKNet backbones \cite{Ding_2024_CVPR}, reporting notable gains on challenging benchmarks. Another line of research targets generalization across object categories and sensing conditions; for instance, FoundationPose \cite{Wen_2024_CVPR} leverages neural implicit representations and large-scale synthetic data to achieve strong cross-domain performance on standardized benchmarks. While these architectural innovations push the state of the forward accuracy, they often introduce increased model complexity, require careful tuning, or depend on specialized training data—factors that can limit their practical adoption and ease of deployment.


\subsection{Pose Refinement}

Pose refinement techniques aim to enhance initial pose estimates through geometric or learning-based optimization. Classical iterative closest point (ICP) methods \cite{chen1992object} minimize point-to-point distances but often struggle in textureless or cluttered scenes. The advent of deep learning shifted the paradigm toward learned refinement: DeepIM \cite{LiIm} pioneered an iterative update scheme using differentiable rendering, while GDR-Net \cite{Wang_2021_GDRN} introduced a geometry-guided direct refinement module within a single forward pass. More recent approaches integrate physical constraints and explicit uncertainty modeling. For example, Repose \cite{Iwase_2021_ICCV} employs deep reinforcement learning to adaptively minimize feature-metric residuals. Coupled Iterative Refinement \cite{lipson2022coupled} proposes a differentiable bidirectional PnP layer (BD-PnP) that jointly refines 2D–3D correspondences and the final pose while dynamically rejecting outliers. In the category-level setting, GeoReF \cite{Zheng_2024_CVPR} addresses intra-class shape variation via geometric alignment using hierarchical semantic (HS) layers and cross-cloud transformation, reporting significant gains over prior refinement baselines.


\subsection{Rotation Representation}
Rotation representation significantly impacts pose estimation performance. Traditional representations include rotation matrices (9D) and Euler angles (3D), but both suffer from discontinuities and constraints (orthogonality/gimbal lock) \cite{zhou2019continuity}. Quaternions (4D) became popular for their compactness and continuity, yet ambiguity (q = -q) complicates loss formulation \cite{chen_projective_2022}. Recent works focus on learning-friendly representations:  \cite{zhou2019continuity} decoupled rotation into two orthogonal axes, avoiding singularities while enabling Gram-Schmidt recovery.  \cite{chen_projective_2022} addressed gradient misalignment in Riemannian manifolds (e.g., SO(3)) by projecting Euclidean gradients onto tangent spaces, improving convergence for quaternion/6D/9D representations. In \cite{Chen_2021_CVPR}, the authors decoupled the rotation representation into two vectors, which are perpendicular to each other, to improve the rotation estimation accuracy. For category-level pose estimation, GPV-Pose \cite{Di_2022_CVPR} integrated geometry-guided point voting and decoupled rotation losses to handle intra-class shape variations. Recent trends also explore implicit representations (e.g., NeRFs for refinement) and equivariant networks \cite{Yuyang_ICCV_2025} for rotation-invariant point cloud features. However, challenges persist in applying these new representations to different methods. 

\subsection{Symmetries in Pose Estimation}
In the field of 6D object pose estimation, handling object symmetry is a core challenge. Existing approaches primarily address it at the levels of loss function design, output representation, and post-processing. Early methods based on correspondence matching often suffered from pose ambiguity due to symmetry. Subsequent research has improved by explicitly modeling symmetry. One category of methods introduces symmetry-aware loss functions during training, such as treating symmetric poses as equivalent labels and computing a minimum distance loss to accommodate multiple valid solutions\cite{Xiang2017}. Another line of work focuses on designing symmetry-invariant intermediate representations, for example, predicting object-local coordinates or symmetry-aware keypoints to avoid ambiguity at the regression stage \cite{wang2019densefusion, peng2018pvnet, Zhao_2023_ICCV}. In recent years, methods have managed symmetry by defining symmetry-robust surface coordinate mappings or aggregating symmetric equivalent solutions through voting mechanisms \cite{pitteri2019object, Hodan_2020_CVPR}. Besides, some methods designed symmetry-invariant loss functions and normalizing the output pose space \cite{Mo_2022_CVPR}. Others proposed a theoretically grounded rotation normalization method that maps all visually equivalent rotations to a canonical representation with multi-regressors. While significant progress has been made for common symmetric objects, efficiently and uniformly modeling symmetry for objects with complex, continuous, or partial symmetry remains an open research direction.

\section{Proposed Method}
While contemporary research predominantly focuses on advancing network architectures for object pose estimation, we identify a fundamental yet overlooked weakness: the reliance on arbitrarily defined, and thus geometrically unstable, object coordinate systems.
This section presents our core innovation: a data-centric optimization paradigm that replaces these ad hoc frames with a canonical, physically grounded coordinate system defined by the object's principal axes of inertia. 
Our method is agnostic to network architectures, offering a plug-and-play preprocessing step that systematically enhances training data to achieve more stable and efficient learning of rotation representations.

\begin{figure*}[htp]
\centering
\includegraphics[width=0.95\textwidth]{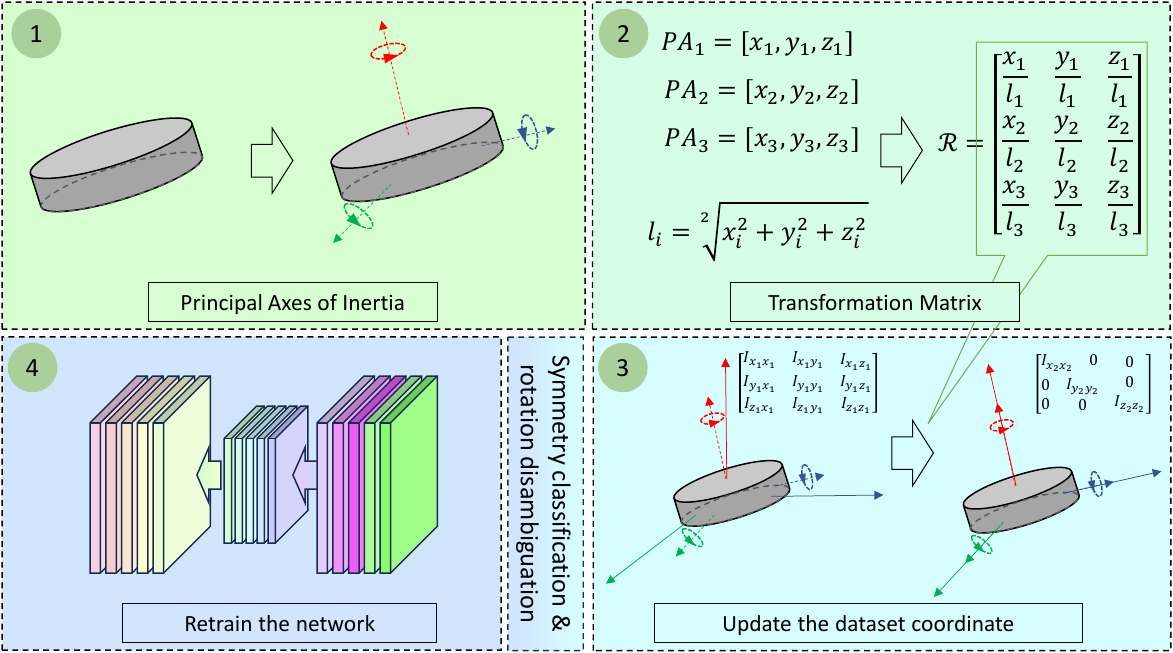}
\caption{\textbf{Overview of the proposed mechanism and its usage.} There are four main steps of our proposed mechanism. First, we calculate the principal axes of the object's inertia, and then we form the transformation rotation using these three principal axes. Third, we access the new object coordinate for the datasets with the transformation rotation. Then, we eliminate the rotation ambiguity with the proposed module if the objects are symmetric.
Finally, we retrain the network using datasets in new coordinates, eliminating rotation ambiguity.}
\label{Fig:over_c_all}
\end{figure*}

The main steps of our proposed $\pi$-Mechanism are shown in Fig. \ref{Fig:over_c_all}. The proposed mechanism comprises four key stages.
First, the principal axes of inertia for the target object are computed to determine its intrinsic geometric orientation. 
Subsequently, a rotational transformation matrix is constructed from these three orthogonal axes. This matrix is then applied to the original dataset, translating all data points into a new, canonical coordinate system defined by the principal axes. 
Finally, the neural network is either retrained from scratch or fine-tuned using the transformed dataset, enabling it to learn features within this normalized and physically meaningful frame of reference. We explain the details of the above steps in the following sections.

\subsection{Geometrically Stable Representation via Principal Axes Alignment}
\subsubsection{Principal Axes}
The core of our geometric optimization lies in replacing the arbitrarily defined object coordinate system with one grounded in the physical properties of the object itself. 
We achieve this by defining the canonical frame based on the object’s principal axes of inertia, which are derived from its mass distribution geometry. 
Formally, for a object 3D model $ \mathcal{M} = \{\textbf{v}_i \in \mathbb{R}^3\}^N_{i=1}$, where $\textbf{v}_i$ means the $i^{th}$ vertex of the 3D model, we first compute the centroid $\textbf{c}=\frac{1}{N}\Sigma_i\textbf{v}_i$, of $ \mathcal{M}$. Then the inertia tensor $\mathcal{I}_{tensor} \in \mathbb{R}^{3\times3}$, which characterizes the mass distribution relative to \textbf{c}, is calculated as:
\begin{equation}
    \mathcal{I}_{tensor}=\sum_{i=1}^N(\|\mathbf{v}_i-\mathbf{c}\|^2\cdot\mathbf{I}-(\mathbf{v}_i-\mathbf{c})(\mathbf{v}_i-\mathbf{c})^\top),
\end{equation}
where $\mathbf{I}$ is the $3 \times 3$ identity matrix. According to the relative theory \cite{1992Eigenvectors, Pagano1995The}, the principal axes are obtained via eigen-decomposition of this symmetric tensor:

\begin{equation}
\label{eq:eigv}
\mathbf{I}_{tensor}\cdot\mathbf{e}_j=\lambda_j\cdot\mathbf{e}_j,\quad j\in\{1,2,3\}.
\end{equation}

The eigenvectors $\mathbf{e}_1, \mathbf{e}_2, \mathbf{e}_3$ define the orthogonal principal axes. These three vectors constitute our intrinsic coordinated frame $\mathcal{F}_{PA} = {\mathbf{e}_1, \mathbf{e}_2, \mathbf{e}_3}$. 

\begin{figure}[htp]
\centering
\includegraphics[width=0.45\textwidth]{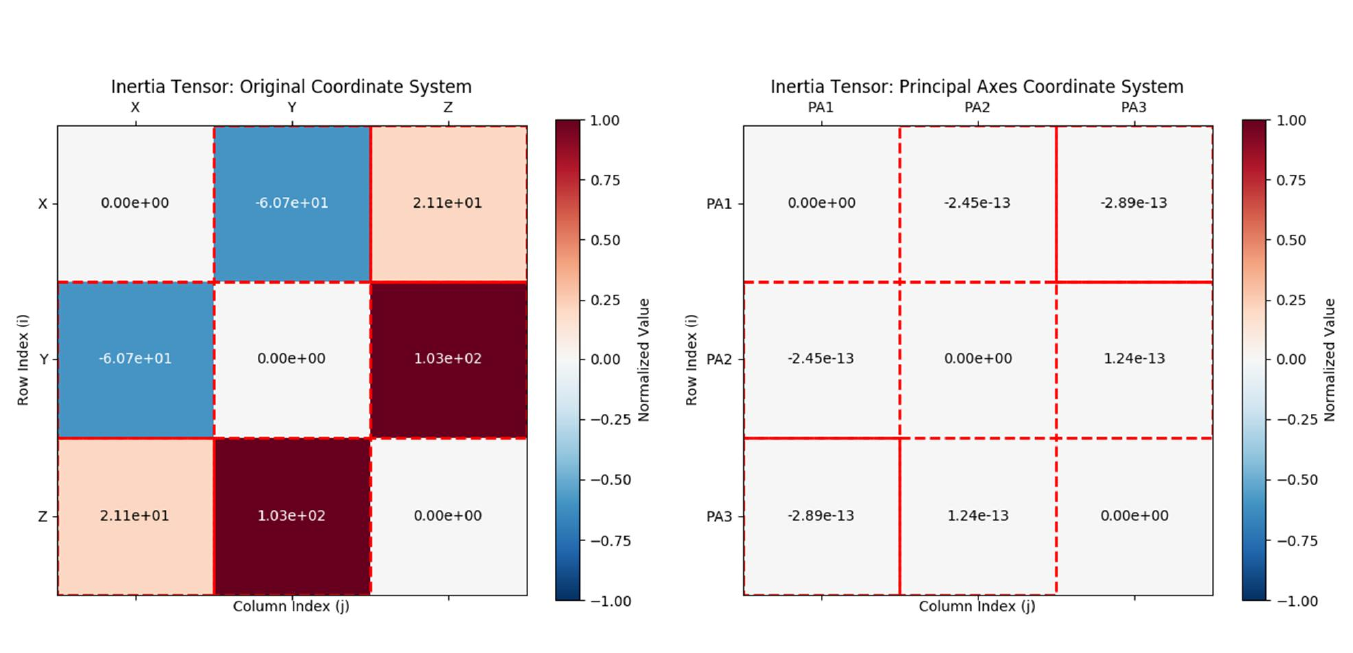}
\caption{\textbf{Physics property of $\pi$-Mechanism}. For original coordinates, the non-diagonal elements are non-zero. For the object in Principal Axes, the non-diagonal elements are close to zero.}
\label{Fig:ori_pai_tensor}
\end{figure}

There are several advantages of our proposed mechanism:
1) rotation stability: the inertia tensor is diagonal in the proposed coordinate $\mathcal{F}_{PA}$, which means rotations around these axes are decoupled and correspond to the object's natural modes of rotation with minimal dynamic disturbation across different axes (as shown in Fig. \ref{Fig:ori_pai_tensor}, the non-diagonal elements are close to zero in $\mathcal{F}_{PA}$). This property translates directly to reduced sensitivity to perturbations in pose estimation. 
2) symmetry invariance: the proposed axes naturally align with the object's intrinsic symmetry axes. ensuring that symmetric transformations (such as rotations by multiples of $180^{\circ}$ for two-fold symmetric objects) leave the axes representation unchanged. Based on this property, we can detect and eliminate rotation ambiguity automatically (as illustrated in Fig. \ref{Fig:sym_reduce_ex}, where symmetric rotations map to identical canonical poses).

\subsubsection{The Alignment Transformation}
\label{subsec:alignment}

Based on the established principal axes as the intrinsic object frame $\mathcal{F}_{PA}$, we now formalize the canonical transformation that aligns any object instance from its original, arbitrary coordinate system $\mathcal{F}_{orig}$ to $\mathcal{F}_{PA}$. Specifically, we apply a pure rotation to the object's coordinate system. The transformation is defined as follows.

Let $\mathbf{V} = [\textbf{v}_1, \textbf{v}_2, \textbf{v}_3] \in \mathbb{R}^{3 \times 3}$  be the matrix whose columns are the three principal axes 
computed in the above section. We construct the alignment rotation matrix $\mathbf{R}_p^o \in SO(3)$ as:
\begin{equation}
    \mathbf{R}_p^o = \mathbf{V},
\end{equation}
which represents the rotation from the original object coordinate system to the principal axes system.

\begin{figure}[htp]
\centering
\includegraphics[width=0.45\textwidth]{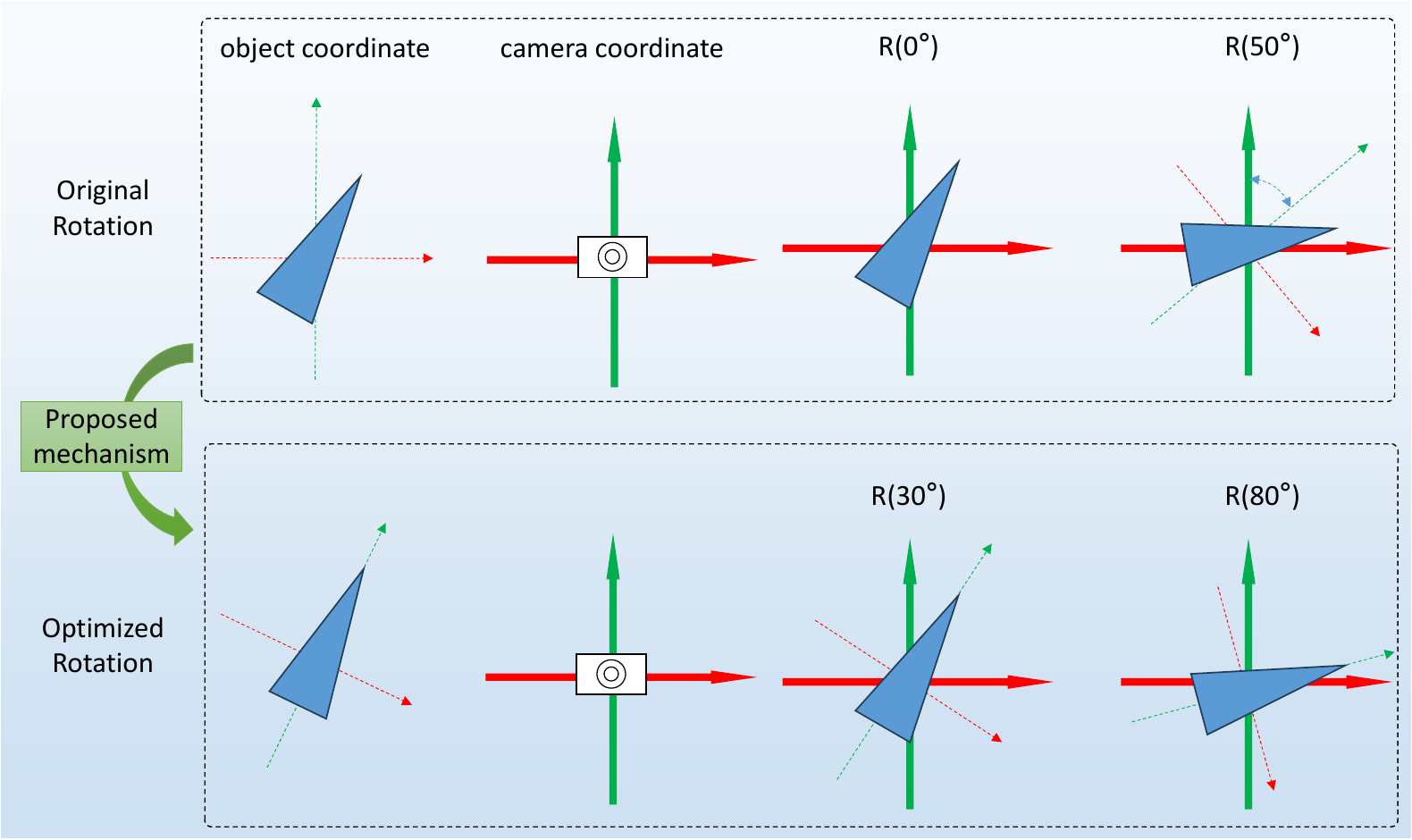}
\caption{\textbf{Illustration of the effect of $\pi$-Mechanism}. The above row shows the geometric relation between the object and its coordinates. When the rotation value is 0, the orientation of the object is not aligned well with the axes of the object coordinate. The second row of images shows the relation when applying our proposed mechanism to the original object relation. The orientation of the object is aligned well with its coordinate. }
\label{Fig:effect_pi}
\end{figure}

Given a ground-truth pose in the original frame, represented as a rotation $\mathbf{R}_{ori}$, the corresponding rotation in the aligned frame is obtained by (shown in Fig. \ref{Fig:effect_pi}): 

\begin{equation}
\mathbf{R}_{\mathrm{PA}}= \mathbf{R}_p^o \cdot \mathbf{R}_{\mathrm{orig}}.
\end{equation} 

Notably, the translation component remains unchanged. It is defined with respect to the object's centroid, which is consistent across both coordinate systems.

This alignment is applied as a preprocessing step to the entire training dataset. For each training sample, the 3D model vertices $\mathcal{M}_\mathrm{orig}$ are canonically oriented via $\mathcal{M}_\mathrm{PA}=\mathbf{R}_p^o\cdot\mathcal{M}_\mathrm{orig}$, and the associated rotation labels are transformed accordingly. During inference, a network predicts a rotation $\mathbf{R}_{\mathrm{pred\to PA}}$ relative to $\{O_{\mathrm{PA}}\}$, which can be converted back to the original frame for
evaluation or application as $\mathbf{R}_{\mathrm{pred}\to\mathrm{orig}}= {\mathbf{R}_p^o}^T \cdot\mathbf{R}_{\mathrm{pred}\to\mathrm{PA}}.$ 
This bi-directional, rotation-only mapping ensures that our geometrically stable representation is seamlessly integrated into the standard pose estimation pipeline.

\subsection{Canonical Reduction of Rotational Ambiguity for Symmetric Objects}
A fundamental property of rigid-body dynamics ensures that any axis of rotational symmetry of an object must coincide with a principal axis of its inertia tensor. This arises because the inertia tensor, as a second-order moment integral of the mass distribution, remains invariant under the symmetry operations defining the object. Consequently, the eigen-decomposition of the inertia tensor naturally reveals and aligns with the object's intrinsic geometric symmetries.
According to this, another pivotal advantage of our approach is its explicit handling of symmetry through physical properties. 
In the following, we will describe how we handle the symmetric object for pose estimation. We first detect whether an object is symmetric. If symmetry is confirmed, classification rules are applied, ambiguous rotation labels are mapped to a canonical space, and the network is trained using these canonical rotations.

\subsubsection{Symmetry Detection \& Classification}
The core premise is that an object is rotationally symmetric if and only if a rotation around a certain axis by some angle $\theta$ results in a shape that is congruent to its original configuration.

Formally, let $\mathcal{P} \in \mathbb{R}^{N \times 3}$ denote the centered object point cloud (with centroid at the origin), and $\mathbf{V'} = [\mathbf{v'}_1, \mathbf{v'}_2, \mathbf{v'}_3] \in \mathbb{R}^{3 \times 3}$ be its matrix of inertial principal axes. The symmetry detection is performed independently for each principal axis $\mathbf{v}_k$ $(k = 1, 2, 3)$ as follows:

\begin{itemize}

    \item \textbf{Angular Space Discretization}: \\
    An angular resolution $\Delta\theta$ (e.g., $5^\circ$) is defined. A sequence of test angles is generated: $\Theta = \{\theta_i \mid \theta_i = i \cdot \Delta\theta, \; i = 1, 2, ..., \lfloor 360/\Delta\theta \rfloor\}$.

    \item \textbf{Rotation and Shape Matching}: \\
    For each $\theta_i \in \Theta$, the rotation matrix $\mathbf{R}(\mathbf{v}_k, \theta_i) \in \text{SO}(3)$ about axis $\mathbf{v}_k$ is computed. The point cloud is rotated to obtain $\mathbf{P}_i' = \mathbf{P} \cdot \mathbf{R}(\mathbf{v}_k, \theta_i)^T$. 
    
    The shape congruence error $d_i$ is then calculated as the bidirectional Chamfer Distance between the original $\mathbf{P}$ and the rotated $\mathbf{P}_i'$:
    \label{eq:chamfer}
\begin{align}
\nonumber
&d_i = D_{CD}(\mathbf{P}, \mathbf{P}_i') = \\
\nonumber
 & \frac{1}{|\mathbf{P}|} \sum_{\mathbf{x} \in \mathbf{P}} 
\min_{\mathbf{y} \in \mathbf{P}_i'} \|\mathbf{x} - 
\mathbf{y}\|_2 \\ &+ \frac{1}{|\mathbf{P}_i'|} 
\sum_{\mathbf{y} \in \mathbf{P}_i'} \min_{\mathbf{x} \in
 \mathbf{P}} \|\mathbf{y} - \mathbf{x}\|_2.
\end{align}

\item \textbf{Symmetry Angle Determination}: \\
    Given a predefined symmetry tolerance threshold $\tau$ (determined by point cloud scale and noise), an angle $\theta_i$ is considered valid if $d_i < \tau$, indicating self-congruence after rotation. The set of all valid angles for axis $\mathbf{v}_k$ is:
    \begin{equation}
        S_k = \{ \theta_i \in \Theta \cup \{0^\circ, 360^\circ\} \mid d_i < \tau \}.
    \end{equation}
    The angles $0^\circ$ and $360^\circ$, representing the identity rotation, are included by default.

\end{itemize}

The rotation symmetry of an object can be automatically detected based on the above calculation. We classify the symmetry of the object as follows.

We first check the symmetry of the object about the first axis of $\mathcal{F}_{PA}$ based on the structure of $S_1$:
    \begin{itemize}
        \item \textbf{Continuous Symmetry}: If $S_1$ contains all angles in $\Theta$ (i.e., $d_i < \tau$ for all $\theta_i$), the object possesses continuous rotational symmetry about this axis (e.g., cylinder, cone). 
        \item \textbf{N-fold Discrete Symmetry}: If $S_1 = \{0^\circ, 360^\circ/N, 2\times360^\circ/N, ..., 360^\circ\}$ for an integer $N > 1$, the object has N-fold discrete rotational symmetry (e.g., a square prism has 4-fold symmetry).
        \item \textbf{2-fold Symmetry}: If $S_1 = \{0^\circ, 180^\circ, 360^\circ\}$, the object has 2-fold (bilateral) symmetry about this axis (e.g., a cuboid rotated about the axis through the center of its faces).
        \item \textbf{No Symmetry}: If $S_1 = \{0^\circ, 360^\circ\}$, the object is asymmetric about this axis.
    \end{itemize}

\begin{figure*}[htp]
\centering
\includegraphics[width=0.9\textwidth]{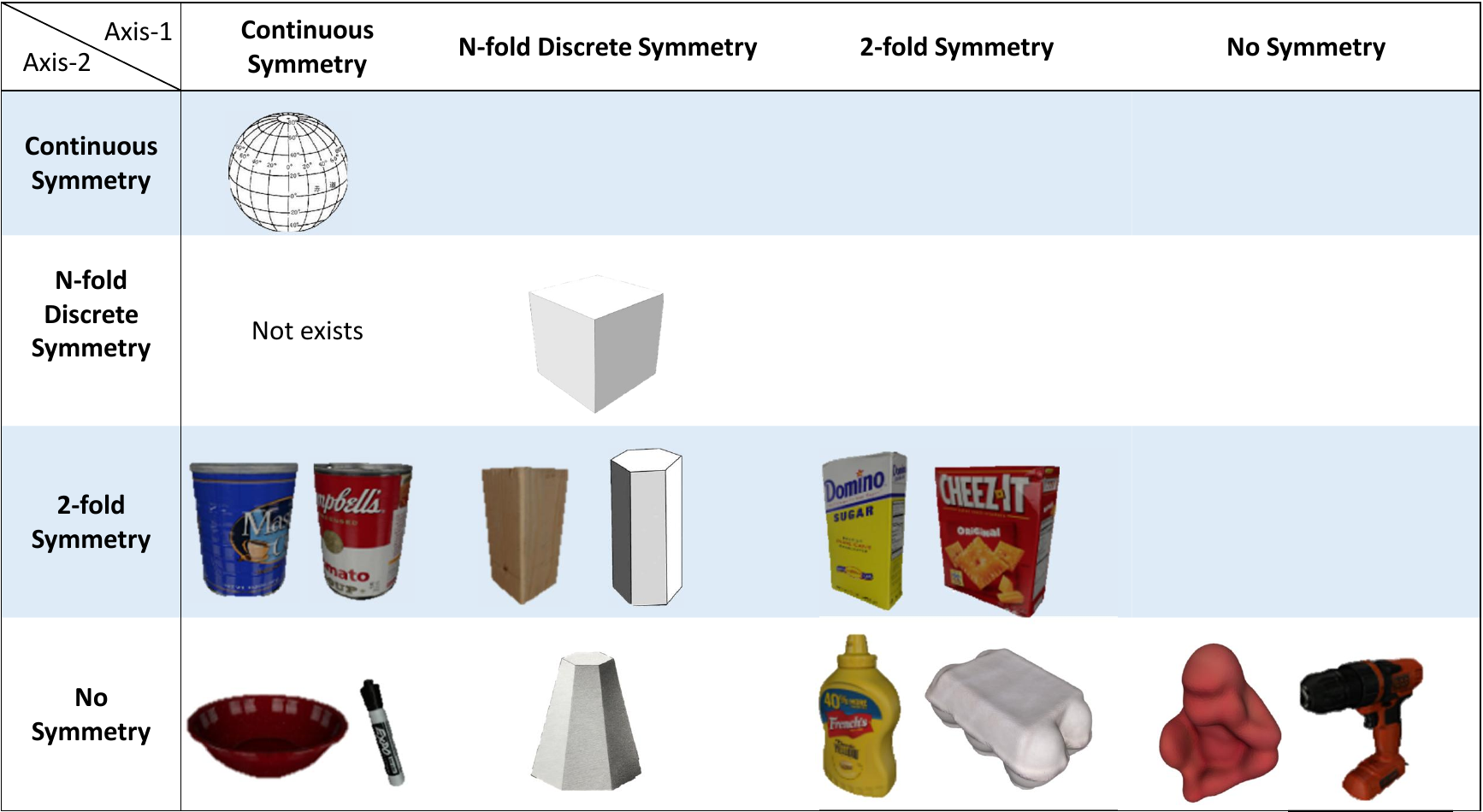}
\caption{\textbf{A Taxonomy of 3D Objects Based on Rotational Symmetry Around Two Principal Axes}. We systematically categorize objects' shapes by their rotational symmetry properties around two independent principal inertial axes. The rows and columns represent the four fundamental symmetry types for a single axis: Continuous, N-fold Discrete, 2-fold, and Asymmetric. Each cell provides one or two canonical 3D object examples (without considering the color texture) exhibiting the corresponding pairwise combination of symmetries. }
\label{Fig:sym_cl}
\end{figure*}

We then check the symmetry of the object about the second axis based on the structure of $S_2$ with the same operation. Without doubt, the symmetry of the object about the second axis can still be classified into four categories: Continuous Symmetry, N-fold Discrete Symmetry, 2-fold Symmetry, and No Symmetry. Then we classify the whole symmetry of an object based on the category combination of those two axes (as shown in Fig. \ref{Fig:sym_cl}). Please note that, due to the geometric constraint \cite{zhou2019continuity}, we do not need to check the third axis.

\subsubsection{Mapping Ambiguous Rotations}

Following the symmetry classification defined in Fig. \ref{Fig:sym_cl}, a critical step is to resolve the inherent rotational ambiguities for symmetric object categories. This process aims to map all physically equivalent poses within a category to a single, unique canonical pose in the dataset. This ensures a bijective mapping between an object's appearance and its pose label, which is essential for stable network training.

\begin{figure}[htp]
\centering
\includegraphics[width=0.45\textwidth]{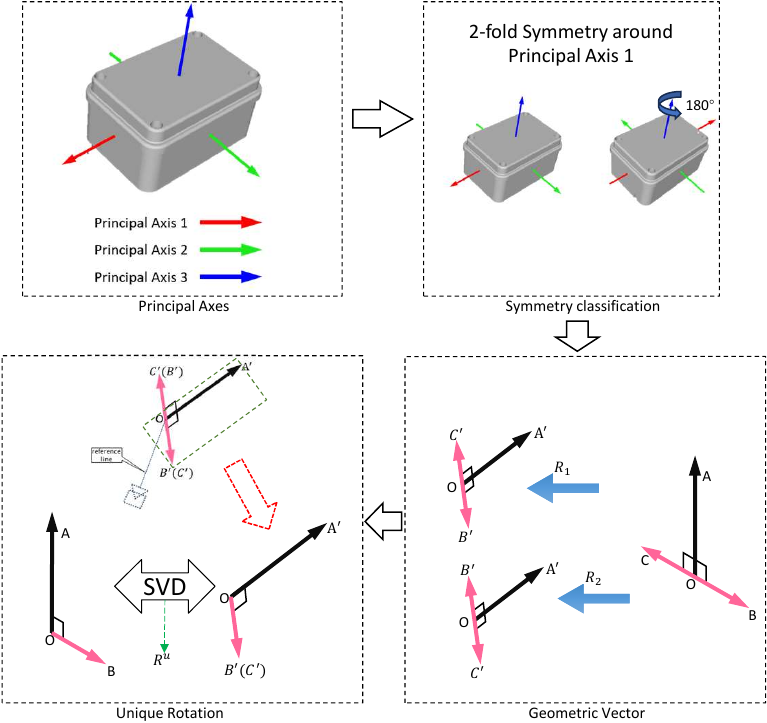}
\caption{\textbf{Pipeline for resolving rotational ambiguity in symmetric objects.} Left top: Computation of the principal axes of inertia from the object’s geometry. Right top: Symmetry type classification (e.g., two-fold symmetry) based on the principal axes. Right bottom: Generation of canonical candidate vectors. Left bottom:  Final unambiguous rotation representation obtained after disambiguation.}
\label{Fig:sym_reduce}
\end{figure}

\begin{figure}[htp]
\centering
\includegraphics[width=0.35\textwidth]{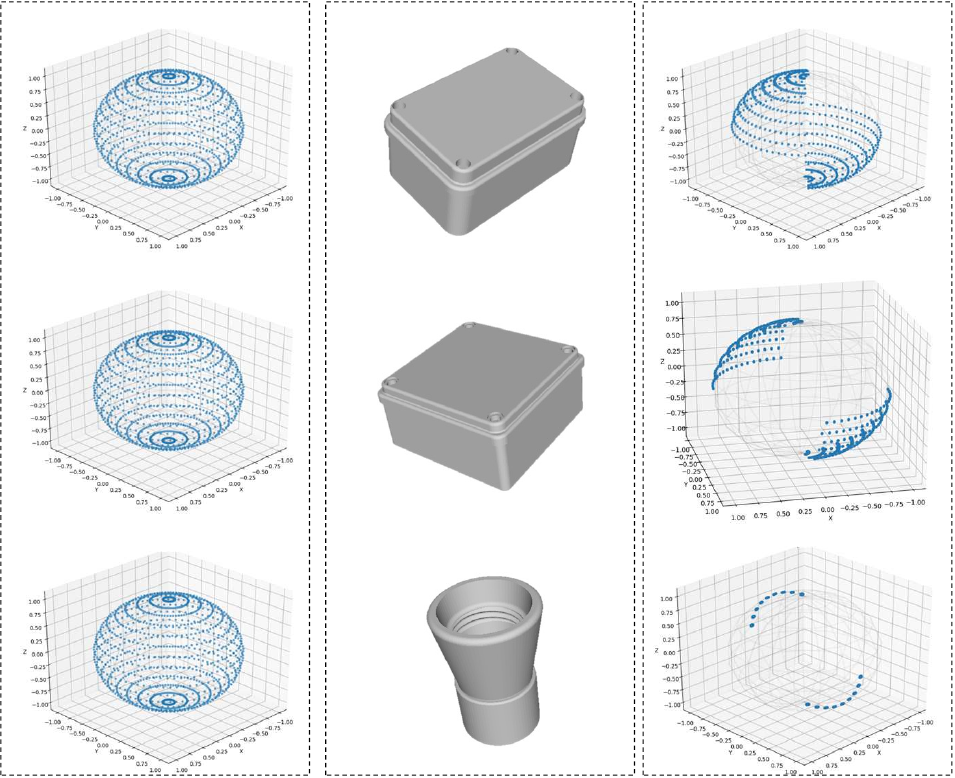}
\caption{\textbf{Illustration of rotation disambiguation for symmetric objects.} Left column: Distribution of rotation viewpoints in the original ambiguous space. Middle column: 3D model of the object in the TLESS dataset (obj29, obj27, obj14), highlighting its intrinsic symmetry. Right column: Unique distribution of rotation viewpoints after ambiguity resolution.}
\label{Fig:sym_reduce_ex}
\end{figure}

Specifically, to systematically address object symmetry at the data-preprocessing stage, we propose a principal axes-based method (shown in Fig. \ref{Fig:sym_reduce}, Fig. \ref{Fig:sym_reduce_ex}). 
We first align the object point cloud to the principal axes frame. Within this canonical frame, the algorithm performs a rotation-congruence test around each of the two principal axes at discrete angular intervals. For each rotated point cloud, the bidirectional Chamfer distance to the original is computed; if this distance falls below a predefined geometric tolerance threshold, the rotation angle is validated as a symmetry angle. The set of all such valid angles for an axis constitutes its rotational ambiguity set, which subsequently classifies the axis as having continuous, N-fold discrete, 2-fold, or no rotational symmetry. For different symmetry classifications, we employ the following strategies to map the ambiguous rotation labels to a unique set.

\begin{figure}[htp]
\centering
\includegraphics[width=0.47\textwidth]{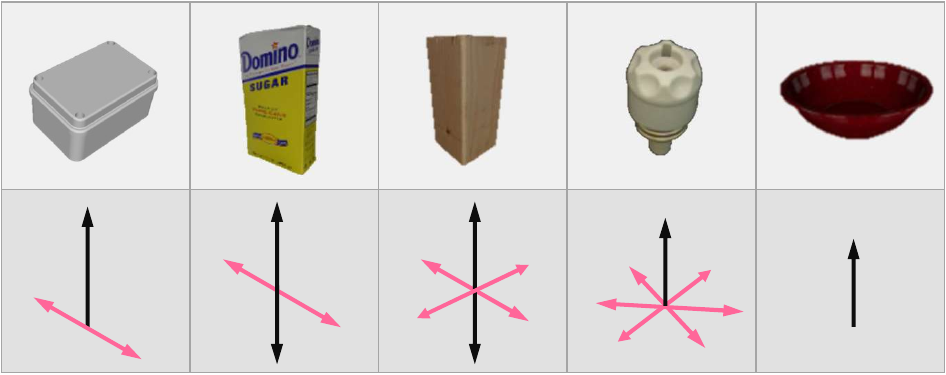}
\caption{\textbf{Illustration of Geometric Vectors for Different Symmetry Types.} }
\label{Fig:sym_vectors}
\end{figure}

In contrast to prior approaches that necessitate additional network branches or specialized loss functions, our proposed solution is straightforward, analytical, and operates directly on the dataset level. As illustrated in Fig. \ref{Fig:sym_vectors}, we design distinct geometric vectors for different symmetry categories according to the following rules: If an object exhibits rotational symmetry only around the first axis (as shown in the last row of Fig. \ref{Fig:sym_cl}), this axis is taken as the primary vector. The plane perpendicular to it is equally divided into N sectors using N vectors, where N corresponds to the fold of rotational symmetry. If a second symmetry axis exists, it is treated as the primary vector, and the same division is applied to its perpendicular plane. Finally, mergeable vectors are combined, resulting in the configuration depicted in Fig. \ref{Fig:sym_vectors}.

Once the geometric vectors are defined, we utilize the rotation labels from datasets of symmetric objects to reorient these vectors accordingly. As demonstrated in Fig. \ref{Fig:sym_reduce_ex}, rotations belonging to the same ambiguous set yield transformed geometric vectors that are indistinguishable in direction—thereby transferring ambiguity from the rotation label (which relates to object appearance) to the vector directions. A reference direction is then established to select a unique vector, which corresponds to a canonical rotation. This process allows the network to be trained exclusively on rotation labels that are free of ambiguity. Please note that any direction can serve as the reference direction, but it should be unique for a specific object within a particular dataset.

The core advantage of this method is rooted in its physical foundation. The inertial principal axes framework transforms the problem from an infinite search in $SO(3)$ to a finite verification of three deterministic, physically-grounded directions. 
It provides an intrinsic and stable object-centric reference frame, guaranteeing consistent symmetry analysis across all instances of the same category. This not only guides the detection process but also enables subsequent disambiguation rules to be based on stable and unique geometric directions. Thus, the process does not merely detect ambiguities but physically constrains them, which makes the rotation disambiguation simple and effective.

Although our method can eliminate the rotation ambiguities of symmetrical objects in the widely used datasets, such as LINEMOD\cite{hinterstoisser2012gradient}, YCB-V\cite{Xiang2017}, and T-LESS\cite{hodan2017t}, it still has some limitations. For example, it cannot collapse the full symmetry group of polyhedra like the octahedron or dodecahedron into a single canonical pose. While one can still use our method to automatically alleviate the rotation ambiguities of polyhedra like the octahedron or dodecahedron, and then manually eliminate the remaining rotation ambiguities.

\section{Experiments}
\noindent In this section, we deploy our $\pi$-Mechanism with different pose estimation methods to show the effectiveness of the proposed mechanism. For all baselines, we use their official codebases and training configurations without any modification (e.g., network architecture, hyperparameters, loss functions, or data augmentation pipelines). Our principal axes alignment (including rotation disambiguation) is applied solely as a preprocessing step to the 3D models and their associated pose annotations in the training and testing sets. All experiments are conducted on a PC with NVIDIA GeForce RTX 4090D GPU.

\subsection{Datasets and Evaluation Metrics}
We evaluate our method on three public benchmarks that span both instance-level and category-level pose estimation, demonstrating its broad applicability. We use rotation errors or ADD-(S) metric for comparison.

\noindent \textbf{LINEMOD} \cite{Hinterstoisser2013} is a widely used instance-level 6D object pose estimation dataset which consists of 13 different objects with significant shape variation. We use the standard split of 15\% for training and 85\% for testing. Please note, for symmetric objects `Egg box' and `Glue' in LINEMOD,  we only use ADD-(S) metric for comparison.

\noindent \textbf{LINEMOD-OCC} \cite{Brachmann2014} is derived from \textbf{LINEMOD} dataset for occlusion scenes. It contains 1214 images, which are annotated for 8 objects under different levels of occlusion.

\noindent \textbf{NOCS-REAL} \cite{wang2019nocs} is the first real-world dataset for category-level 6D object pose estimation. The training set has 4300 real images of 7 scenes with 6 categories. For each category, there are 3 unique instances. In the testing set, there are 2750 real images spread across 6 scenes of the same 6 categories as the training set. In each test scene, there are about 5 objects, which makes the dataset cluttered and challenging.

\subsection{Toy Experiments on Synthetic LINEMOD}
We first validate the proposed method on a synthetic LINEMOD dataset. This dataset is rendered from the original LINEMOD CAD models \cite{Hinterstoisser2013} by applying random rotations to each object while keeping the translation fixed at zero. Please note that samples vary only in rotation, and are free of noise or background clutter to isolate the effect of our geometric preprocessing on rotation estimation.

We employ two representative backbones for rotation regression: PointNet \cite{Qi_2017_CVPR} and SI-Mamba \cite{Bahri_2025_CVPR}. PointNet is a foundational architecture for direct point cloud processing, widely adopted in subsequent research \cite{Qi_2018_CVPR, dgcnn}. SI-Mamba integrates the hierarchical feature extraction of PointNet++ \cite{qi2017pointnetplusplus} with the efficient sequence modeling of Mamba \cite{mamba, mamba2}, representing a recent advance in network design.

The results in Table \ref{tab:abla-syn} show that our principal axes alignment improves rotation prediction accuracy by 3.3\% for PointNet and 4.0\% for SI-Mamba. This consistent gain across two distinct architectures demonstrates that the proposed data-centric optimization enhances rotation estimation inherently and is invariant to the choice of network backbone.


\begin{table*}[t]
\caption{\textbf{Rotation error for different methods on the synthetic LINEMOD dataset.} The best result is bolded.}
\label{tab:abla-syn}
\centering
\resizebox{2\columnwidth}{!}
{
\begin{tabular}{lcccccccccccc}
\hline
Method &  Ape &  Bench Vise  & Camera & Can & Cat & Driller  & Glue & Hole Puncher & Iron & Lamp & Phone& Average\\
\hline
PointNet \cite{Qi_2017_CVPR}& 12.28 &	12.49 &	11.76 &	13.71	&13.11 &  11.30 &	\textbf{10.72} &\textbf{10.23} & 11.62 & 10.47 & 11.09& 11.71\\

PointNet +  $\pi$-Mechanism & \textbf{12.12} & \textbf{10.30} & \textbf{10.27} & \textbf{13.16} & \textbf{12.51} &  \textbf{11.26} & 11.41 & 12.34 & \textbf{11.42} & \textbf{10.01}  & \textbf{9.71} & \textbf{11.32} \\
\hline

Si-Mamba \cite{Bahri_2025_CVPR}
& 10.85 &  11.04  & 10.42 & \textbf{11.34} & 11.22 & 11.06 & 10.13 & 12.55 & \textbf{10.17} & 11.57 & 11.34 & 11.06 \\
Si-Mamba +$\pi$-Mechanism & \textbf{10.48} & \textbf{10.58} &\textbf{10.05} &{11.94} &\textbf{10.26}&\textbf{10.49}&\textbf{9.86}& \textbf{11.46} & {10.67} & \textbf{10.25} & \textbf{10.72} & \textbf{10.61}\\

\hline
\end{tabular}
}
\end{table*}

\subsection{Experiments on LINEMOD}
Building upon the demonstrated improvement in rotation estimation within a controlled synthetic environment, we now transition to evaluating our method under more challenging, real-world conditions. To this end, we employ the standard LINEMOD benchmark and integrate our preprocessing with two established pose estimation frameworks: G2L-Net \cite{Chen_2020_CVPR} and GDR-Net \cite{Wang_2021_GDRN}. This experiment is designed to validate the practical effectiveness and general applicability of our geometric alignment approach.

\noindent \textbf{Comparison on G2L-Net.} 
G2L-Net is a real-time 6D pose estimation framework that processes RGB-D images in a divide-and-conquer fashion. It begins by extracting coarse object point clouds via 2D detection, followed by a translation localization network that performs 3D segmentation and predicts translation. The points are then transformed into a canonical coordinate system, where a rotation localization network estimates an initial rotation along with a residual refinement. By leveraging viewpoint-aware embedding features, the method achieves state-of-the-art accuracy on benchmarks such as LINEMOD while maintaining high frame rates. We retrain G2L-Net using the official setup from \cite{Chen_2020_CVPR}.

As shown in Table \ref{tab:abls-rgdb}, integrating our $\pi$-Mechanism with G2L-Net yields consistent improvement across all objects in the LINEMOD dataset. Notably, the mean rotation error decreases from 9.61 to 7.09—a relative reduction of 26.2\%. This significant enhancement demonstrates that the proposed geometric alignment effectively stabilizes the rotation estimation, complementing the strong baseline performance of G2L-Net.

\noindent \textbf{Comparison on GDR-Net} 
GDR-Net is an end-to-end framework that directly regresses 6D object pose from a single RGB image. Its key innovation lies in leveraging dense 2D–3D correspondences, structured in a patch-based representation, and enhancing feature discrimination through a surface region attention (SRA) mechanism. We retrain GDR-Net on the LINEMOD dataset using its original loss function based on the $R_{6D}$ representation \cite{zhou2019continuity}, while integrating our proposed $\pi$-Mechanism as the only modification.

The results demonstrate that our $\pi$-Mechanism consistently reduces the rotation error across categories. As shown in Table~\ref{tab:abls-rgdb}, the mean rotation error decreases from 2.01 to 1.83—an 8.96\% relative improvement. This confirms that even for a strong correspondence-based baseline, canonical alignment provides a measurable gain in rotation estimation accuracy.

\begin{table*}[t]
\caption{\textbf{Rotation error for different methods on LINMOD dataset.} The best result is bolded.}
\label{tab:abls-rgdb}
\centering
\resizebox{2\columnwidth}{!}
{
\begin{tabular}{lcccccccccccccc}
\hline
Method &  Ape &  Bench Vise  & Camera & Can & Cat & Driller & Duck  & Hole Puncher & Iron & Lamp & Phone& Average\\
\hline
G2L \cite{Chen_2020_CVPR}& {7.37} & 15.34 & 5.99 & 8.96 & {6.03} & 13.04 & 6.84  & 5.13 & 17.20 & 12.54 & 7.24 & 9.61\\
G2L +  $\pi$-Mechanism & \textbf{6.25} &	\textbf{11.94} &	\textbf{5.83} &	\textbf{6.30}	&\textbf{5.23} &	\textbf{9.37} &	\textbf{6.58} &	\textbf{3.94}&	\textbf{7.37}	&\textbf{9.79}&	\textbf{5.44}&	\textbf{7.09}\\

\hline

GDR-Net\cite{Wang_2021_GDRN}
& 2.11 & 1.85 &1.81 &1.82 &2.02&2.02&1.98 & 1.96 & 2.33 & 1.87 & 2.30 & 2.01\\


GDR-Net+$\pi$-Mechanism & \textbf{1.83} & \textbf{1.82} &\textbf{1.67} &\textbf{1.73} &\textbf{1.80}&\textbf{1.83}&\textbf{1.83} & \textbf{1.88} & \textbf{2.04} & \textbf{1.68} & \textbf{2.19} & \textbf{1.83}\\

\hline
\end{tabular}
}
\end{table*}

\begin{figure}[ht!]
\centering
\includegraphics[width=0.49\textwidth]{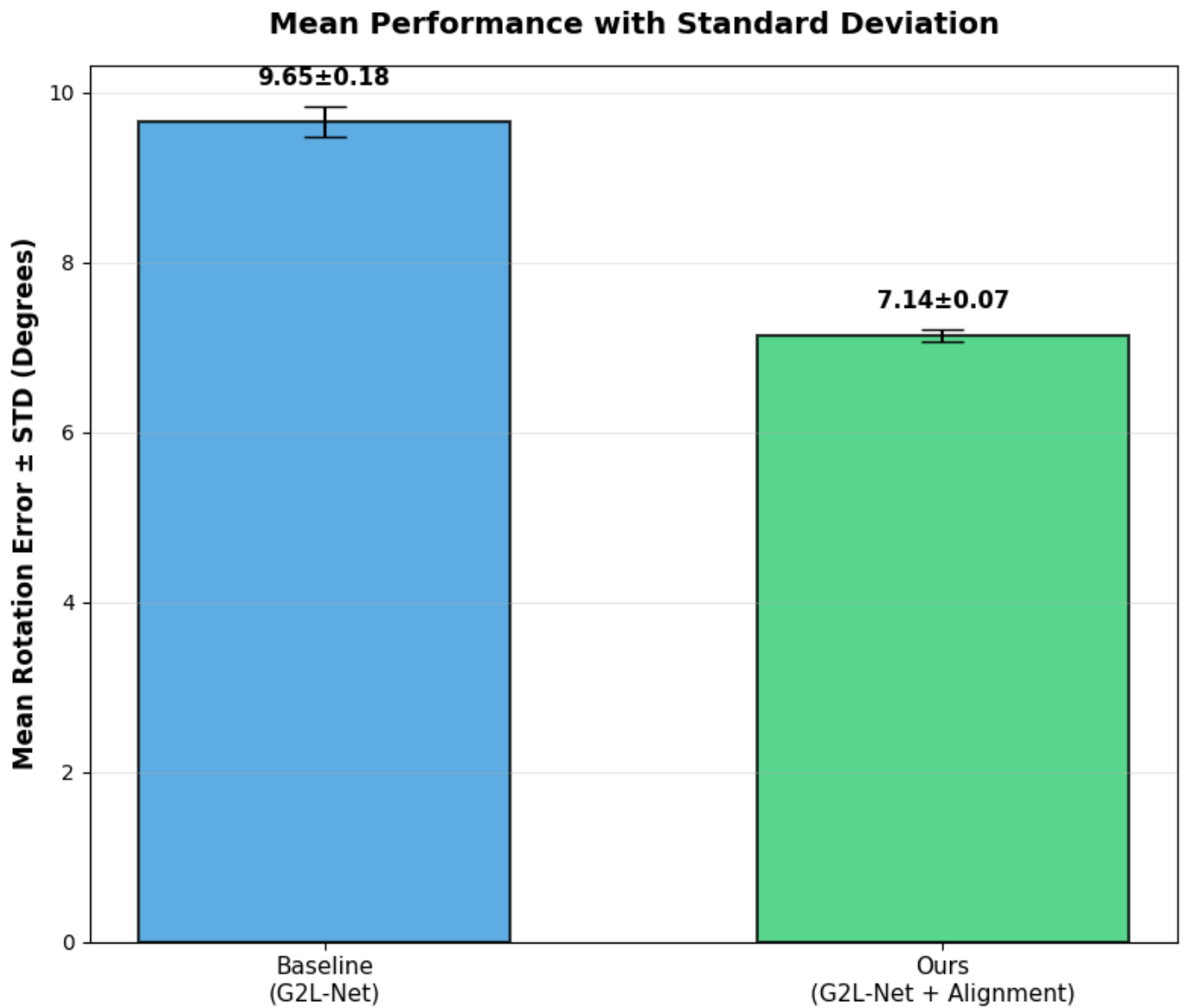}
\caption{\textbf{Mean rotation errors with standard deviation (SD).} The left bar is the mean rotation error and SD of the baseline method; The right bar is the mean rotation error and SD of the proposed mechanism.}
\label{Fig:loss_times}
\end{figure}

\begin{figure*}[ht!]
\begin{center}
\begin{tabular}{cccc}
{{\includegraphics[width=0.225\textwidth]{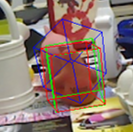}}}&
{{\includegraphics[width=0.225\textwidth]{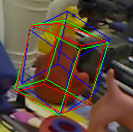}}}&{{\includegraphics[width=0.225\textwidth]{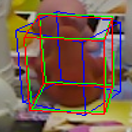}}}&{{\includegraphics[width=0.225\textwidth]{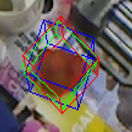}}}\\
{{\includegraphics[width=0.225\textwidth]{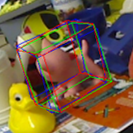}}}&{{\includegraphics[width=0.225\textwidth]{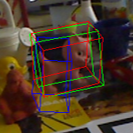}}}&{{\includegraphics[width=0.225\textwidth]{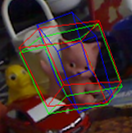}}}&{{\includegraphics[width=0.225\textwidth]{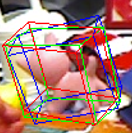}}}\\
{{\includegraphics[width=0.225\textwidth]{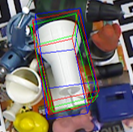}}}&{{\includegraphics[width=0.225\textwidth]{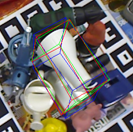}}}&{{\includegraphics[width=0.225\textwidth]{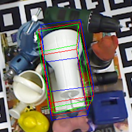}}}&{{\includegraphics[width=0.225\textwidth]{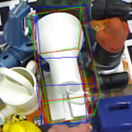}}}\\
{{\includegraphics[width=0.225\textwidth]{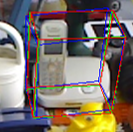}}}&{{\includegraphics[width=0.225\textwidth]{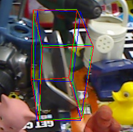}}}&{{\includegraphics[width=0.225\textwidth]{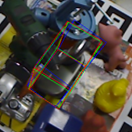}}}&{{\includegraphics[width=0.225\textwidth]{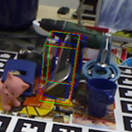}}}\\

\end{tabular}
\end{center}
\vspace{-10pt}
\hspace{-10pt}
\caption{\textbf{The Quantitative results for the LINEMOD dataset.} The green boxes are the ground truth. The red boxes are our results. The blue boxes are the results of the baseline.}
\label{Fig:ins-quality-line}
\end{figure*}

Fig. \ref{Fig:ins-quality-line} provides a qualitative comparison of challenging samples from the LINEMOD dataset. The visualization employs a consistent color scheme, with green representing the ground-truth, blue representing the baseline method, and red representing our approach. 
It can be observed that our method (red) consistently produces bounding boxes that are more accurately aligned with the ground-truth (green) geometry, particularly under the texture-less scene(Row 1). In contrast, the baseline predictions (blue) frequently exhibit noticeable rotational errors or instability. These visual results corroborate our quantitative findings, demonstrating that alignment to a stable, intrinsic coordinate system enables more robust and accurate rotation estimation under diverse challenging conditions.

To rigorously evaluate the robustness and reproducibility of our method beyond a single training run, we conducted a statistical experiment using G2L-Net \cite{Chen_2021_CVPR} as the baseline on the LINEMOD dataset. We performed five independent trials, each with random weight initialization and data shuffling. For each trial, we trained two models: the original G2L-Net (Baseline) and G2L-Net trained with our principal axes-aligned data (Ours). Fig. \ref{Fig:loss_times} presents the box plots of the mean rotation error across all 5 times for both configurations.

The results are statistically compelling. Our method achieves a mean rotation error of 7.14, with a standard deviation of 0.07. In contrast, the baseline yields a higher mean of 9.65 and a notably higher standard deviation of 0.18. This demonstrates that our approach not only decreases the rotation error by 26\% but also significantly reduces the variance in outcomes by 61\%. The reduced variance indicates that the trained models are more stable and less dependent on random initialization when learning a rotation representation anchored in the geometrically stable principal axes frame.

\subsection{Experiments on LINEMOD-OCC}
To further evaluate robustness under severe occlusion, we test the proposed mechanism on the challenging LINEMOD-OCC dataset. We adopt LC (Linear Covariance) \cite{Liu_2023_ICCV} as the baseline, which introduces a loss function enabling direct end-to-end 6D pose regression from RGB images. By formulating the pose in Lie algebra, LC operates in the tangent space of the SO(3) rotation manifold, providing a geometrically consistent and stable learning signal that improves both accuracy and convergence for direct regression networks.

The results are summarized in Table \ref{tab:abla-lc}. With the integration of our $\pi$-Mechanism, pose accuracy measured by the ADD metric improves by 3.4\% on average across all object categories. A notable case is the `Ape' object, whose accuracy rises from 44.44\% to 48.97\%, corresponding to a 10.2\% relative gain. These results confirm that our geometric alignment enhances pose estimation even under highly occluded conditions.

\begin{table*}[h]
    \centering
    \caption{Results on dataset \textbf{LINEMOD-OCCLUDE} for method GDR-LC \cite{Liu_2023_ICCV} in ADD(-S)-0.1d metric (the higher is better). The best result is bolded. For symmetric objects `Egg Box' and `Glue', the ADD-S scores are reported.}
    \begin{tabular}{c|c|c|c|c|c|c|c|c|c}
    \hline
    Methods &Ape & Can & Cat & Driller & Duck & Egg box & Glue & Hole Puncher & Average \\ \hline
LC  & 44.44\%   & 89.06\%   & 49.87\%   & 87.81\%   & 56.08\%   & 62.81\%   & 68.88\%  & 72.89\%  & 66.69\% \\ 
LC+$\pi$-M  & \textbf{48.97\%}   & \textbf{90.22\%}   & \textbf{51.39\%}   & \textbf{89.46\%}   & \textbf{57.31\%}   & 65.37\%   & 70.86\%  & \textbf{76.61\%}  & \textbf{68.99\%} \\ 
    
    \hline
    \end{tabular}
        \label{tab:abla-lc}
\end{table*}

\begin{figure*}[ht!]
\begin{center}
\begin{tabular}{cccc}
{{\includegraphics[width=0.225\textwidth]{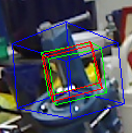}}}&
{{\includegraphics[width=0.225\textwidth]{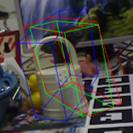}}}&
{{\includegraphics[width=0.225\textwidth]{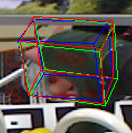}}}&
{{\includegraphics[width=0.225\textwidth]{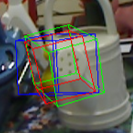}}}\\
{{\includegraphics[width=0.225\textwidth]{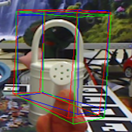}}}&
{{\includegraphics[width=0.225\textwidth]{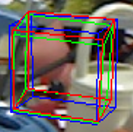}}}&
{{\includegraphics[width=0.225\textwidth]{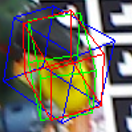}}}&
{{\includegraphics[width=0.225\textwidth]{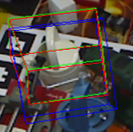}}}\\

\end{tabular}
\end{center}
\vspace{-10pt}
\hspace{-10pt}
\caption{\textbf{The Quantitative results for LINEMOD-OCC dataset.} The green boxes are the ground truth. The red boxes are our results. The blue boxes are the results of the baseline.}
\label{Fig:ins-quality-occ}
\end{figure*}

Fig. \ref{Fig:ins-quality-occ} presents a qualitative comparison on the challenging LINEMOD-OCC dataset with LC\cite{Liu_2023_ICCV} as the baseline. We follow the scheme: ground-truth (green), LC baseline prediction (blue), and our method’s prediction (red). As shown, the LC baseline (blue) frequently produces bounding boxes with large rotational errors when the target object is partially obscured. In contrast, our method (red), leveraging the stable representation aligned with the object’s principal axes, consistently estimates poses that are much closer to the ground truth (green), even under severe occlusion. This visual evidence directly supports the quantitative improvements reported in TABLE \ref{tab:abla-lc}, demonstrating that our data-centric optimization significantly enhances robustness in real-world occlusion scenarios.

\subsection{Experiments on Symmetric Objects}
To test the effectiveness of our $\pi$-Mechanism on symmetric objects, we pick up some typical symmetric objects from the widely used TLESS dataset (shown in Fig. \ref{Fig:sym_reduce_ex}). We then report the rotation errors of these objects in TABLE \ref{tab:tless_sym}. From the table, we can see that the rotation errors drop dramatically with the optimized dataset.

\begin{table}[t]
\caption{\textbf{Rotation error for different methods on TLESS symmetric objects.} The best result is bolded. }
\label{tab:tless_sym}
\centering
{
\begin{tabular}{lccc}
\hline
Method &  obj 14 &  obj 27  & obj 29  \\
\hline
G2L \cite{Chen_2020_CVPR} & 31.51 & 71.70  & 108.83  \\
G2L + $\pi$-Mechanism & \textbf{2.64} & \textbf{10.11} &\textbf{17.79} \\

\hline
\end{tabular}
}
\end{table}

\subsection{Category-level Tasks}
Unlike instance-level tasks, category-level pose estimation is challenged by significant intra-class variation in shape and appearance. To validate the applicability of our proposed mechanism in this setting, we conduct experiments on the NOCS dataset \cite{wang2019nocs}.

\begin{figure}[h]
\centering
\includegraphics[width=0.35\textwidth]{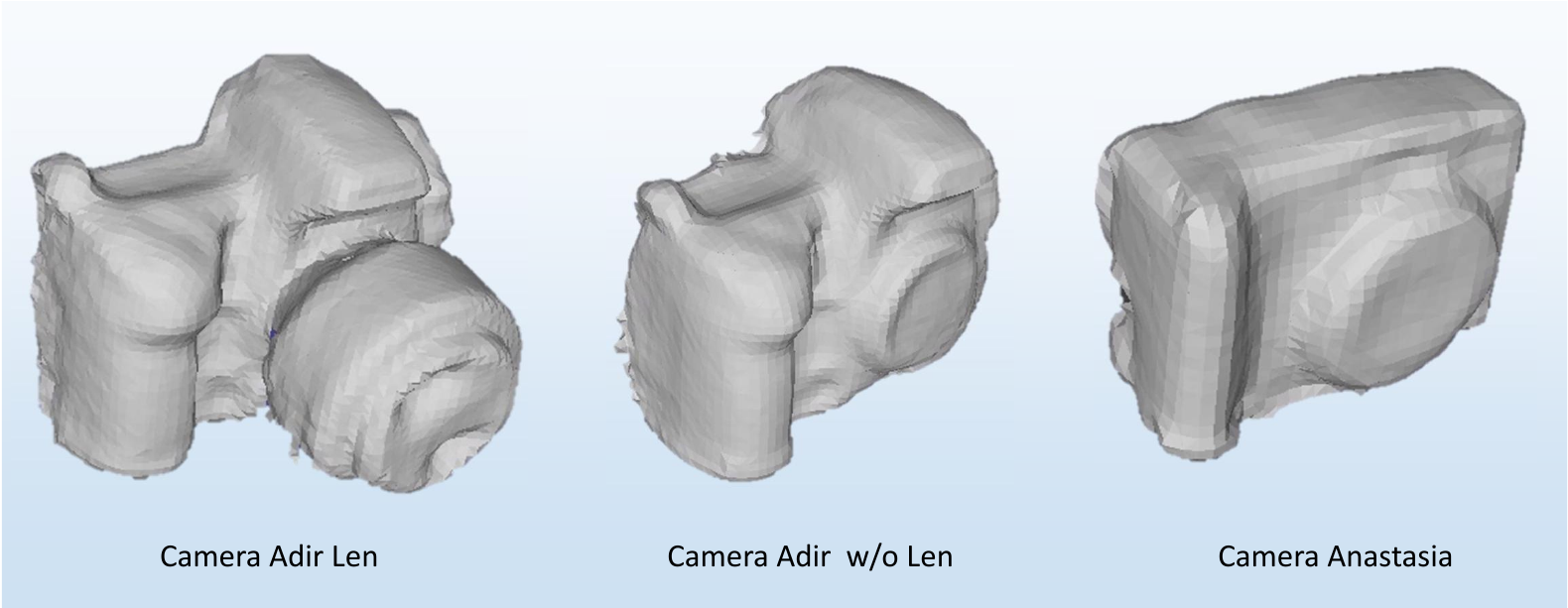}
\caption{`Camera' category with different instances}
\label{Fig:camins}
\end{figure}

A key adaptation for the category-level task is the computation of the Principal Axes of Inertia. Since objects within a category can have substantially different geometries, we compute the principal axes for each object instance independently. For example, the "camera" category in NOCS contains three distinct instances with markedly different shapes, as illustrated in Fig. \ref{Fig:camins}. During training, the respective principal axes for each instance are derived and used for canonical alignment. This instance-specific alignment ensures that our geometric normalization remains effective despite intra-class shape variation.

For the category-level pose estimation experiments, we adopt the FS-Net pipeline as our baseline. Following the evaluation protocol in \cite{zhou2019continuity}, we use the geodesic error to assess different rotation representations. The quantitative results are presented in Table \ref{tab:fsrots}.

As shown in Table \ref{tab:fsrots}, our proposed $\pi$-Mechanism achieves the best performance for rotation estimation. It reduces the mean rotation error from 11.27 to 8.92—a relative improvement of 19.6\% over the baseline. This result confirms that our geometry-driven alignment generalizes effectively to the category-level setting, where handling significant intra-class shape variation is critical.

\begin{table*}
\centering
\caption{\textbf{Rotation error comparison.} Results on dataset \textbf{NOCS-REAL} for method FS-Net \cite{Chen_2021_CVPR} in rotation prediction error. The best result is bolded.}
\label{tab:fsrots}
\begin{threeparttable}

\begin{tabular}{clccccccc}
\hline
 &Category & Bottle & Bowl & Can & Camera & Laptop & Mug &  Average \\ \hline

\multirow{2}{*}{different representation}&FS-Net&{5.71} & {3.85} & {3.97} & {21.59} & {8.00} & {13.77} & {9.48}\\
&FS-Net+$\pi$-Mechanism & \textbf{5.34} & \textbf{3.64} & \textbf{3.58} & \textbf{20.99} & \textbf{7.57} & \textbf{12.40} & \textbf{8.92}\\

 \hline
\end{tabular}

\end{threeparttable} 
  
\end{table*}

\subsection{Convergence and Learning Dynamics}
To move beyond empirical metrics and understand the fundamental mechanism behind our performance gains, we investigate how principal axes alignment reshapes the underlying learning problem. We analyze three interconnected aspects: optimization convergence, feature representation, and loss landscape geometry, all evaluated using the same network architecture where the only variable is the coordinate representation (original vs. aligned).

\begin{figure}[h]
\centering
\includegraphics[width=0.45\textwidth]{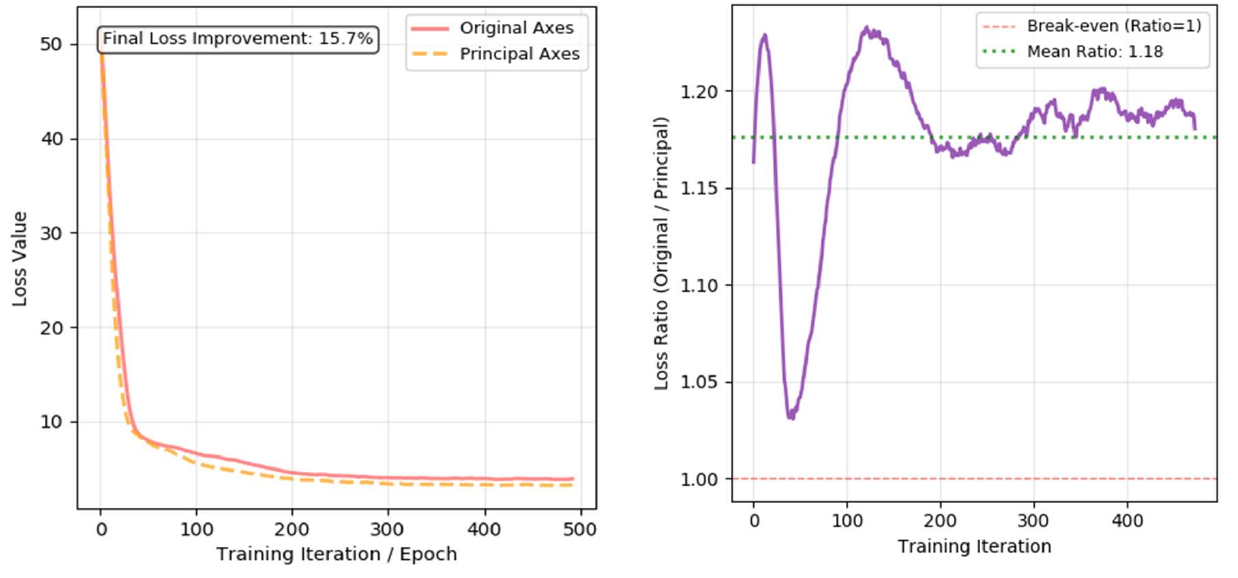}
\caption{\textbf{Loss value comparison}. Left: the loss value comparison between original datasets and datasets in Principal Axes coordinates. Right: the loss ratio between original and principal over training iterations.}
\label{Fig:loss_reduce}
\end{figure}

We monitor the training dynamics. As shown in Fig. \ref{Fig:loss_reduce}, the rotation loss for networks trained on our aligned data converges to a lower plateau compared to training on original data. This indicates that regressing rotations relative to the stable principal axes constitutes a simpler and more well-posed learning objective.

To summarize, by aligning the training data to the geometrically stable principal axes, we reparameterize the rotation estimation task into an intrinsically "easier" space. This reformulation results in a lower loss landscape, provides more stable gradients, and enables the network—without any architectural changes—to learn more coherent and robust feature representations. This fundamental improvement in learning dynamics directly explains the consistent accuracy gains observed across diverse networks and datasets.

\section{Conclusion}
In this work, we have introduced a fundamental shift in perspective for 6D object pose estimation: from modifying deep networks to optimizing their geometric input. Our proposed method aligns object coordinates with their intrinsic Principal Axes, establishing a canonical, physics-informed rotation representation that is inherently more stable. Extensive experimentation—from synthetic poses to occluded real-world scenes, and from instance-level to category-level tasks—demonstrates that this data-centric, model-agnostic optimization consistently and significantly improves accuracy across diverse architectures. 
The core of our contribution is threefold. First, we provide a geometrically principled solution that directly addresses the instability of arbitrary object coordinates, thereby enhancing robustness against noise and occlusion. Second, we propose an analytical yet straightforward method to resolve ambiguous rotation labels for symmetric objects. Third, we establish a plug-and-play paradigm that decouples geometric optimization from network design, enabling immediate performance gains without retraining or modifying existing models. 
This work not only presents an effective mechanism but also opens a new, promising direction for advancing pose estimation through geometric and data-centric reasoning, potentially influencing other vision tasks reliant on stable 3D representations.

\newpage
\bibliographystyle{IEEEtran}
\bibliography{egbib}

 
\vspace{11pt}

\vspace{11pt}

\vfill

\end{document}